\documentclass[conference]{IEEEtran}
\IEEEoverridecommandlockouts
\usepackage{cite}
\usepackage{amsmath,amssymb,amsfonts}
\usepackage{algorithmic}
\usepackage{graphicx}
\usepackage{textcomp}
\usepackage{xcolor}
\def\BibTeX{{\rm B\kern-.05em{\sc i\kern-.025em b}\kern-.08em
    T\kern-.1667em\lower.7ex\hbox{E}\kern-.125emX}}
    
\makeatletter
\newcommand{\linebreakand}{%
  \end{@IEEEauthorhalign}
  \hfill\mbox{}\par
  \mbox{}\hfill\begin{@IEEEauthorhalign}
}
\makeatother

\usepackage{enumitem}
\usepackage{url}
\usepackage[hidelinks]{hyperref}
\hypersetup{
  colorlinks,
  citecolor=blue,
  linkcolor=red,
  urlcolor=blue}
\usepackage{ragged2e}
\usepackage{tabularx}
\usepackage{multirow}
\usepackage{float}
\usepackage{booktabs}
\usepackage{changes}

\usepackage{hhline}

\begin{document}

\title{Towards Practical Explainability with Cluster Descriptors\\

\thanks{This work is supported by Fujitsu Research of America, Inc.}
}

\author{1\textsuperscript{st} \IEEEauthorblockN{Xiaoyuan Liu}
\IEEEauthorblockA{\textit{Fujitsu Research of America, Inc.}\\
Sunnyvale, CA, USA \\
xliu@fujitsu.com}
\and
\IEEEauthorblockN{2\textsuperscript{nd} Ilya Tyagin}
\IEEEauthorblockA{\textit{University of Delaware}\\
Newark, DE, USA \\
tyagin@udel.edu}
\and
\IEEEauthorblockN{3\textsuperscript{rd} Hayato Ushijima-Mwesigwa}
\IEEEauthorblockA{\textit{Fujitsu Research of America, Inc.}\\
Sunnyvale, CA, USA \\
hayato@fujitsu.com}
\linebreakand
\IEEEauthorblockN{4\textsuperscript{th} Indradeep Ghosh}
\IEEEauthorblockA{\textit{Fujitsu Research of America, Inc.}\\
Sunnyvale, CA, USA \\
ighosh@fujitsu.com}
\and
\IEEEauthorblockN{5\textsuperscript{th} Ilya Safro}
\IEEEauthorblockA{\textit{University of Delaware}\\
Newark, DE, USA \\
isafro@udel.edu}
}

\maketitle    

\begin{abstract}
With the rapid development of machine learning, improving its explainability has become a crucial research goal. We study the problem of making the clusters more explainable by investigating the cluster descriptors. Given a set of objects $S$, a clustering of these objects $\pi$, and a set of tags $T$ that have not participated in the clustering algorithm. Each object in $S$ is associated with a subset of $T$. The goal is to find a representative set of tags for each cluster, referred to as the cluster descriptors, with the constraint that these descriptors we find are pairwise disjoint, and the total size of all the descriptors is minimized. In general, this problem is NP-hard. We propose a novel explainability model that reinforces the previous models in such a way that tags that do not contribute to explainability and do not sufficiently distinguish between clusters are not added to the optimal descriptors. The proposed model is formulated as a quadratic unconstrained binary optimization problem which makes it suitable for solving on modern optimization hardware accelerators. We experimentally demonstrate how a proposed explainability model can be solved on specialized hardware for accelerating combinatorial optimization, the Fujitsu Digital Annealer, and use real-life Twitter and PubMed datasets for use cases. Reproducibility materials: \href{https://anonymous.4open.science/r/explainability-cluster-descriptor-9307}{Link}
\end{abstract}


\section{Introduction}\label{sec:intro}
Machine learning (ML) methods and artificial intelligence (AI) technologies bring growing impact to our daily life, across all domains, from healthcare to urban planning. As a result, the need to make the obtained results more explainable is continuously growing, and improving the explainability is extremely important for many applications. Results of unsupervised ML techniques are often hard to interpret, especially in a post-hoc analysis. Unsupervised clustering is not an exception.

While most existing literature focuses on performing clustering and finding an explanation simultaneously using the same dataset, Davidson et al.~\cite{davidson2018cluster} first introduced the cluster description problem in which they explored the idea of taking a partitioned dataset $X$  found by some clustering algorithm, and explaining its result using another dataset $Y$. For instance, in our experiments with the Twitter dataset~\cite{davidson2018cluster}, $X$ is a matrix of user reposting behavior in Twitter, i.e., $X_{i,j}$ is the number of times user $i$ was retweeted by user $j$. We are given the result found by spectral clustering, which partitions the users into clusters. The dataset $Y$ is a matrix of users' hashtag usage, i.e., $Y_{i,j}$ is the number of times user $i$ tweets with hashtag $j$. For another example, we experiment with PubMed, the dataset of scientific abstracts and metadata, for which matrix $X$ is not strictly required for clustering. The clustering is naturally performed based on the medical subject heading terms (MeSH terms) that are manually curated and added to the document metadata. Matrix $Y$ represents a document-term matrix i.e., $Y_{i,j}$ is the number of times a term $j$ occurs in document $i$. 
Overall, there exist many motivating examples in which we may need to restrict some features from participating in clustering but then use them to analyze the clusters. One important example is sensitive and protected features that exhibit a major problem in modern data mining. These features are not supposed to participate in clustering but if they still clearly distinguish the obtained clusters, it could be a sign of a wrong clustering model or hyperparameters' choice. Another use case is the third party proprietary data that cannot inform clustering but can be used to analyze its results.

\begin{figure*}[!htbp]
	\centering
	\includegraphics[width=\linewidth]{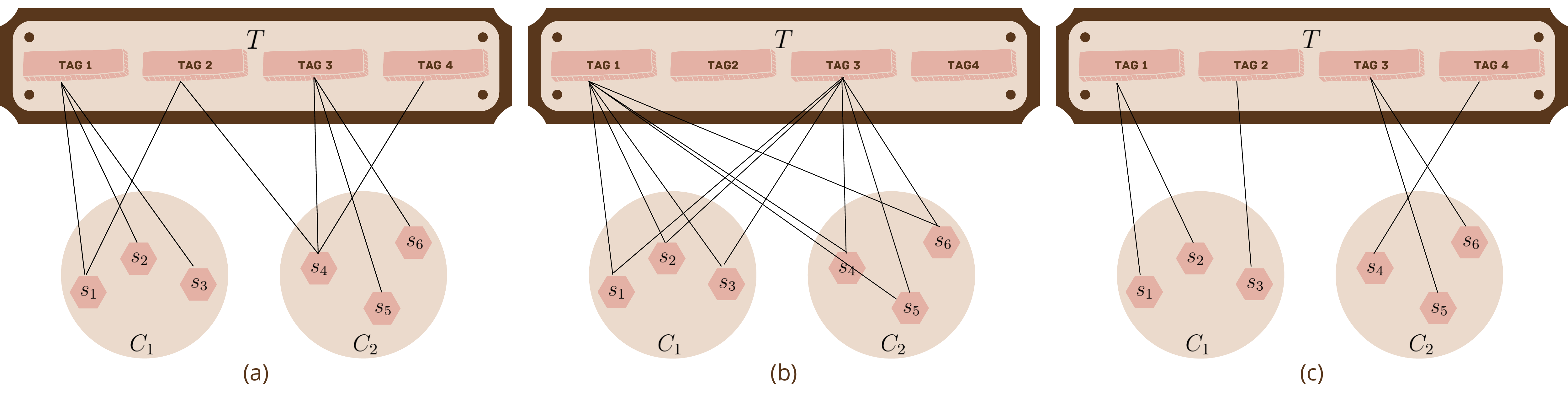}
	\caption{(a) A toy example with 6 objects $\{s_i\}_{i=1}^6$, and their partition into 2 clusters $C_1$ and $C_2$. Tag set $T$ contains 4 tags. Each object $s_i$ is associated a subset $t_i \subset T$. In this example, $t_{1} = \{\texttt{TAG 1}, \texttt{TAG 2}\}$, $t_{2} = \{\texttt{TAG 1}\}$, $t_{3} = \{\texttt{TAG 1}\}$, $t_{4} = \{\texttt{TAG 2}, \texttt{TAG 3}, \texttt{TAG 4}\}$, $t_{5} = \{\texttt{TAG 3}\}$, $t_{6} = \{\texttt{TAG 3}\}$. One solution for DTDM is $T_1 = \{\texttt{TAG 1}\}, T_2 = \{\texttt{TAG 3}\}$. (b) Optimal solution for DTDM and MinConCD is $T_1 = \{\texttt{TAG 1}\}, T_2 = \{\texttt{TAG 3}\}$, however both tags are generic and do not reveal specific information for each cluster, reducing the usefulness of the descriptors. (c) If select $T_1 = \{\texttt{TAG 1}, \texttt{TAG 2}\}, T_2 = \{\texttt{TAG 3}, \texttt{TAG 4}\}$ as descriptors, they explain each cluster better albeit using more tags.}\label{fig:tag}
\end{figure*}

Let $S$ denote the set of $n$ objects, $S = \{s_i\}_{i=1}^n$, and we are given a clustering $\pi$ that partitions $S$ into $k$ disjoint clusters $\{C_{\ell}\}_{\ell=1}^k$. We are also given a set of tags $T$. Each object $s_i\in S$ is associated with a subset $t_i \subseteq T$ of tags, $1\leq i \leq n$. For each cluster, we define  $T_{\ell}\subseteq T$,  the \textbf{descriptor} of cluster $C_{\ell}$. An object $s_i$ in cluster $C_{\ell}$, is said to be \textbf{covered} by the descriptor $T_{\ell}$, if $T_{\ell}$ contains at least one of the tags in $t_i$, the tag set associated with $s_i$. The goal of cluster description problem is to find $k$ pairwise disjoint descriptors, one descriptor  for each cluster, such that all objects are covered and the total number of tags used in the descriptors is minimized. This optimization problem is refered to as the disjoint tag descriptor minimization (DTDM) problem. Figure \ref{fig:tag} (a) depicts a toy example with 6 objects and a partition of 2 clusters.

However, the constraint that requires all objects to be covered can sometimes be too strict, making even finding a feasible solution very challenging. To address this issue, Davidson et al. also discussed several variants of DTDM. One such variant is referred to as the ``cover-or-forget'' version, which relaxes the constraint that all objects must be covered and allows ignoring some of the objects. Recently, Sambaturu et al.~\cite{sambaturu2020efficient},  have pointed out that the ``cover-or-forget'' variant does resolve the feasibility issue but might have highly unbalanced coverage, i.e., for some clusters, the descriptors can cover most of the objects, but for some cluster, the descriptor might cover very few objects. The authors proposed to address this issue by adding constraints on the number of objects needed to be covered for each cluster, i.e., introduce additional parameter $M_{\ell}$ for each cluster $C_{\ell}$, such that the number of objects that are covered must be at least $M_{\ell}$. The extended cluster description formulation with more balanced coverage guarantees for all clusters is referred to as the minimum constrained cluster description (MinConCD) problem. The DTDM is therefore a special case of MinConCD when $M_{\ell} = |C_{\ell}|$ for $1 \leq \ell \leq k$. However, we argue that this model has another issue. Consider the following scenario: in many datasets, we might have several tags that cover most of the objects. For example, it has been noticed multiple times that depending on the partially extracted information, the Twitter hashtags may follow Zipf or approximate power laws~\cite{chen2020scaling,cunha2011analyzing}. If we only focus on minimizing the total number of tags while satisfying the coverage constraints, the most frequently used tags will be selected preferably. Figure \ref{fig:tag} (b) shows such scenario, while the optimal solution uses only 2 tags. These tags might generally describe the entire dataset, they do not reveal specific information about the cluster they are assigned to. The descriptors of the clusters will therefore be similar to each other, reducing the usefulness of the results. In contrast, we wish to select the tags that reveal the information of each cluster albeit potentially using more tags, as in the scenario displayed in Figure \ref{fig:tag} (c).

\emph{To address this issue, we propose to further reinforce the explainability formulations of MinConCD by adding an additional metric to the objective function which will fundamentally change the outcomes.} The novel metric, which we call \textbf{tag modularity (TM)}, is inspired by the notion of modularity \cite{newman2004finding} in network science. Modularity is a measure of the graph structure which quantifies the strengths of the division of a graph into clusters. A clustering with high modularity tends to have dense connections between the nodes within clusters but sparse connections between nodes in different clusters. Quantitatively, the modularity on graphs measures a deviation of node clusters from a random model. Maximizing the modularity naturally punishes trivial clusterings which are practically useless.  Intuitively, by considering both the total number of tags used for the descriptors and the connections between the tags and the objects, we can better identify the cluster-specific tags and filter out the general tags, making the descriptor more interpretable and thus improving the explainability of the descriptors.

Both DTDM and MinConCD are challenging combinatorial optimization problems. Davidson et al. \cite{davidson2018cluster} showed that even for two clusters, finding only a feasible solution of DTDM is intractable not to mention the optimality. Sambaturu et al.~\cite{sambaturu2020efficient} showed that if the coverage constraints in MinConCD for each cluster must be met, then unless $P = NP$, for any $\rho \geq 1$, there is no polynomial-time algorithm that can approximate the objective function value with a factor of $\rho$. Typically, in order to tackle such challenging optimization models, the problems are formulated in terms of integer linear programming (ILP), as proposed in the works of Davidson et al. and Sambaturu et al. Then, to tackle the computational intractability of such problems, relaxations on the constraints or approximation algorithms are applied.

We explore the way to deal with the hardness of such problems from the optimization accelerator systems' angle whose recent advent  opens new opportunities for the data mining community. As we are approaching the end of Moore's law \cite{schaller1997moore} due to the physical limitations, major efforts from the academic and industry research are directed towards developing novel hardware, specifically designed for the combinatorial optimization. Examples of such special-purpose hardware include adiabatic quantum computers  \cite{johnson2011quantum}
, complementary metal-oxide-semiconductor (CMOS) annealers  \cite{aramon2019physics} and 
coherent Ising machines  \cite{inagaki2016coherent}. 
Gate-based quantum computers can also be used to solve such optimization problems  \cite{ushijima2021multilevel,farhi2014quantum,liu2021layer}. Although these emerging technologies exhibit different hardware designs and architecture, they can all be unified by the mathematical framework of the Ising model or equivalently, quadratic unconstrained binary optimization (QUBO) problem.

The focus of developing specialized hardware to solve QUBO stems from its potential applications and computation challenges~\cite{kochenberger2014unconstrained}. QUBO model is notable for modeling a wide range of problems in physics~\cite{barahona1988application}, optimization problems on graphs~\cite{boros1991max,pardalos1994maximum}, network science~\cite{shaydulin2018community,shaydulin2019network,ushijima2017graph} and many more. While many of these problems such as spectral clustering and community detection in social network analysis can be naturally formulated as QUBO, the applicability of QUBO can be significantly extended by a reformulation that can convert a constrained problem into an equivalent QUBO problem, i.e., including quadratic penalties in the objective function as an alternative to imposing the constraints explicitly. Many NP-hard combinatorial optimization problems can be easily and efficiently formulated as  QUBO~\cite{glover2018tutorial,lucas2014ising}. In addition, QUBO can also be used as a subroutine for local search approaches to solve large scale problems \cite{shaydulin2019hybrid,liu2022leveraging} in which a global problem is decomposed into many small ones. The model's ubiquity and widespread usefulness make the development of specialized hardware for solving it a fruitful academic and industrial endeavor. \emph{In this paper, in addition to the novel explainability model, we explore the QUBO formulation and applicability of specialized hardware for accelerated solution.}

\textbf{Our Contribution:}
The major goals central to this work are as follows:
\begin{enumerate}
\item We extend the MinConCD problem formulation in \cite{sambaturu2020efficient} and propose a QUBO formulation of a new model, the tag modularity cluster descriptor (TMCD) problem.
\item To improve the explainability of the descriptors, we propose a novel metric, tag modularity, which aims to generate cluster-specific tags for each cluster and drop the generic tags that do not contribute to the practical explainability of clusters.
\item We explore the applicability of specialized hardware accelerator Fujitsu Digital annealer to solve the computationally challenging model.
\item We demonstrate the efficacy of our proposed model by showing two applications: (1) a real-life Twitter dataset in which hashtags are used to explain clusters of users; (2) explaining the clustering of biomedical documents and topic modeling with the Pubmed dataset.
\end{enumerate}

\section{Background}\label{sec:background}
\subsection{Cluster Description Problem}
For the completeness of the paper, we first briefly introduce the MinConCD problem \cite{sambaturu2020efficient}. Given a set of objects $S$, its $k$ disjoint clusters $C_{\ell}$ and a tag set $T$ (as defined in Section \ref{sec:intro}). For each tag $j \in T$ and cluster $C_{\ell}$, $1 \leq \ell \leq k$, we introduce binary variables 
\[
    x_{\ell}(j)= 
\begin{cases}
    1,& \text{if tag $j$ is assigned to the descriptor $T_{\ell}$ of $C_{\ell}$,}\\
    0,& \text{otherwise}.
\end{cases}
\]
We also define a binary variable $z(i)$ for each object $s_i \in S$,
\[
z(i)=\begin{cases}
    1, & \text{if object $s_i\in S$ is covered,}\\
    0, & \text{otherwise}.
\end{cases}
\]
Define parameters $M_{\ell}$ for each cluster $C_{\ell}, 1 \leq \ell \leq k$, to indicate the number of objects need to be covered for each cluster. For each descriptor $T_{\ell}$, let
\[V(T_{\ell}) = \{s_i \in C_{\ell} \mid t_i \cap T_{\ell} \not = \varnothing\}
\]
be the subset of objects in $C_{\ell}$ covered by the descriptor $T_{\ell}$, $1 \leq \ell \leq k$. The goal of MinConCD is to find a subset $T_{\ell} \subseteq T$ of tags for each cluster $C_{\ell}, 1 \leq \ell \leq k$, such that: 

\begin{enumerate}
\item The descriptors $\{T_{\ell}\}_{\ell=1}^k$ are pairwise disjoint.
\item $\sum_{\ell=1}^k |T_{\ell}|$ is minimized.
\item For each cluster $C_{\ell}$, $|V(T_{\ell})| \geq M_{\ell}$.
\end{enumerate}
The ILP formulation of MinConCD is then given as follows: 
\begin{align}
\text{(ILP)}\ & \min \  \sum_{\ell=1}^k\sum_{j\in T} x_{\ell}(j) \\
\text{s.t.}\ &\forall \ell, \forall s_i \in C_{\ell} : \sum_{j\in t_i} x_{\ell}(j) \geq z(i), \\
&\forall \ell: \sum_{s_i\in C_{\ell}} z(i) \geq M_{\ell}, \\
&\forall j: \sum_{\ell=1}^k x_{\ell}(j) \leq 1, \\
&\forall j, \forall \ell: x_{\ell}(j) \in \{0, 1\}, \ \forall i: z(i) \in \{0, 1\}.
\end{align}

\subsection{Quadratic Unconstrained Binary Optimization}
\label{sec:background1}

The quadratic unconstrained binary optimization (QUBO) problem is a common mathematical abstraction in combinatorial optimization that has wide range of applications~\cite{kochenberger2014unconstrained}. The QUBO problem is defined as follows:
\begin{equation}
\begin{aligned}\label{eq:qubo}
\min \ x^T Q x \\
\text{s.t.} \ x \in B,
\end{aligned}
\end{equation}
where $B$ is the binary set $\{0, 1\}^n$ or $\{-1, 1\}^n$ and $Q$ is a $n\times n$ symmetric square matrix of coefficients. Many combinatorial optimization problems are formulated as binary integer programming (IP). Such IPs can be converted to QUBO by standard techniques \cite{lucas2014ising,glover2018tutorial}. The linear constraints in IP are converted to quadratic penalty terms and added to the objective functions. The quadratic penalties are introduced in such a way that if the constraints are satisfied, their contribution to the objective function is zero; and if some constraint is not satisfied, its contribution to the objective function is positive. QUBO is computationally equivalent to the Ising model, an important model in physics. Due to this close connection, QUBO becomes a central problem class for adiabatic quantum computing and specialized hardware accelerators.

\subsection{Digital Annealer}
\label{sec:background2}

The emergence of novel computational devices such as digital annealers, quantum computers, and CMOS annealers presents new opportunities for hybrid optimization algorithms. Recently, Coffrin et al.~\cite{coffrin2019evaluating} proposed the idea of Ising processing unit (IPU) as a computational abstraction for reasoning with these emerging devices. As IPU technology matures, it could become a valuable co-processor in hybrid optimization and ML.

Fujitsu's Digital Annealer (DA) is a hardware accelerator for solving fully connected QUBO problems (i.e., the values of $Q_{ij}$ are nonzero for all $i, j$ in Equation (\ref{eq:qubo})). 
Internally, the hardware runs a modified version of a Markov Chain Monte Carlo (MCMC) method. The strength of the hardware lies mainly in the massively parallel implementation of the MCMC algorithm. At each MCMC step, each bit is considered to be flipped in parallel. Furthermore, in traditional MCMC methods on fully connected QUBO's, the time it takes to accept a bit flip is usually longer than the time taken to reject. This is mainly because accepting a bit flip requires updating the effective fields of all bits. Given that the DA makes these updates in parallel, the time for each step is constant regardless of accepting a bit flip or not. This greatly improves the time to solve. 

In this work, we use the third generation DA \cite{digitalannealer} 
which is a hybrid software-hardware configuration. In the third generation DA, the software component first determines feasible regions in the solution space which are then subsequently analyzed by the hardware. The third generation DA supports up to 100,000 binary variables. This capacity allows using such devices for the mid-  to large-scale optimization tasks in data mining. The user is able to add inequality constraints and special constraints such as 1-hot and 2-way 1-hot constraints that the software layer utilizes for determining potentially feasible regions.

\subsection{Related Work}
The topic of explainable AI has recently attracted major attention. Clustering, as a commonly used unsupervised ML technique, however, is hard to explain and interpret. In the context of explaining clustering results, most literature focuses on simultaneously performing clustering and explanation procedures. The conceptual clustering~\cite{fisher1987knowledge,gennari1989models}, for example, attempts to find the clustering and explain it using the same features and therefore is restricted to a single set of categorical variables. Predictive clustering~\cite{langley1996elements} uses the features to find a clustering and a descriptive function for prediction. Davidson et al. \cite{davidson2018cluster} explore the idea of taking an existing clustering found by dataset $X$ and explaining it using another dataset $Y$. Notably, only $X$ and not $Y$ is used to find the clustering, and therefore, this is not a semi-supervised scenario. The authors present and express the problem as an ILP and solve it using MATLAB ILP solver. In recent work, Sambaturu et al. \cite{sambaturu2020efficient} point out that for many datasets, the exact cover formulation in \cite{davidson2018cluster} often does not have a feasible solution, and the ``cover-or-forget'' variation does give a solution but might have unbalanced coverage. To address these issues, they introduce the MinConCD problem, which has simultaneous coverage guarantees on all the clusters. They present an ILP formulation and design a randomized algorithm for MinConCD based on rounding a linear programming (LP) solution.

\section{Tag modularity cluster descriptor}

Given object set $S$ and tag set $T$, we first construct a graph $G = (V, E)$, with 
\[
V = S \cup T,
\]
and 
\[
E = \{(s_i, j) \mid \forall s_i\in S, \forall j\in T\}.\]
An example of such graph is demonstrated in Figure \ref{fig:tag} (a). The original modularity metric \cite{newman2004finding} is a measure of the structure of graphs that quantifies the strengths of division of a graph into clusters. A clustering with high modularity has dense connections between the nodes within clusters but sparse connections between nodes in different clusters. It computes the fraction of the edges that are within the cluster minus the expected fraction if edges are distributed at random. The modularity of a clustered graph is defined as follows:
\begin{equation*}
Q = \frac{1}{2|E|} \sum_{v,w\in V}\left[A_{vw} - \frac{k_vk_w}{2|E|}\right]\delta(c_v, c_w),
\end{equation*}
where $A$ is the adjacency matrix, i.e., $A_{vw} = 1$ if $(v, w)\in E$ and 0 otherwise. The $k_v$ denotes the degree of node $v\in V$. Let $c_v$ denote the membership of node $v$ in the clustering, namely, $c_v = \ell$ if node $v$ is assigned to cluster $C_{\ell}$. Here, $\delta$ is the Kronecker delta, i.e., $\delta(c_v, c_w) = 1$ if $c_v = c_w$ and 0 otherwise.

The original modularity metric is a sum over all pairs of nodes in the graph. However, note that in our model, the graph has two types of nodes, the object nodes $S$ and the tag nodes $T$, i.e.,  the graph is bipartite. In addition, the clustering of the object nodes $S$ is given. Therefore, instead of summing up over all pairs of nodes, we propose a new metric which we call \textbf{tag modularity (TM)} that only sums up over the tag nodes. Here, the tags are disconnected by construction, i.e., $A_{vw} = 0, \forall v, w\in T$. Formally defined, the tag modularity is given by:
\begin{equation*}
\text{TM} = \sum_{v,w\in T}\frac{k_vk_w}{2|E|}\delta(c_v, c_w).
\end{equation*}
Note that, in the original modularity metric, the larger modularity is, the connections within the clusters are denser. Here, we drop the negative sign, so the smaller TM is, the better. With this novel metric TM, we propose another model, the \textbf{tag modularity cluster descriptor (TMCD)} problem.

Given a set of objects $S$, its $k$ disjoint clusters $C_{\ell}$ and a tag set $T$. The goal of TMCD is to find $k$ disjoint descriptors, that minimize the total number of the selected tags and the tag modularity of the given graph, while guaranteeing the percentage of the objects that are covered. The quadratic programming formulation (QP) is given as follows:
\begin{align}
\text{(QP)}\ & \min \ \sum_{\ell=1}^k\sum_{j\in T} x_{\ell}(j) + P \sum_{\ell=1}^k\sum_{v,w\in T}\frac{k_vk_w}{2|E|}x_{\ell}(v)x_{\ell}(w)\\
\text{s.t.} \ &\forall \ell, \forall s_i \in C_{\ell} : \sum_{j\in t_i} x_{\ell}(j) \geq z(i),\label{eq:constraint1} \\
&\forall \ell: \sum_{s_i\in C_{\ell}} z(i) \geq M_{\ell}, \label{eq:constraint2}\\
&\forall j: \sum_{\ell=1}^k x_{\ell}(j) \leq 1, \label{eq:constraint3}\\
&\forall j, \forall \ell: x_{\ell}(j) \in \{0, 1\}, \ \forall i: z(i) \in \{0, 1\},
\end{align}
where $P \geq 0$ is a parameter, and MinConCD is a special case of TMCD when $P = 0$.

The QP formulation of TMCD can then be relaxed and converted to a QUBO using standard techniques~\cite{lucas2014ising,glover2018tutorial}. For constraints (\ref{eq:constraint1}), we introduce $m_{1,i} = \lceil\log_2|t_i|\rceil$ slack binary variables $\{y_{1,i,b}\}_{b=1}^{m_{i,1}}$ to convert the inequality constraints to equality constraints:
\begin{equation}
\begin{aligned}
\forall \ell, \forall s_i \in C_{\ell}: & \ z(i) - \sum_{j\in t_i} x_{\ell}(j) + \sum_{b=1}^{m_{1,i}-1} 2^{b-1}y_{1,i,b} \\
&\quad + (|t_i|+1-2^{m_{1,i}-1})y_{1,i,m_{1,i}}= 0. \label{eq:constraint4}
\end{aligned}
\end{equation}
For constraints (\ref{eq:constraint2}), we introduce $m_{2,\ell} = \lceil\log_2(|C_{\ell}|- M_{\ell})\rceil$ slack binary variables $\{y_{2,\ell,b}\}_{b=1}^{m_{2,\ell}}$ to convert the inequality constraints to equality constraints:
\begin{equation}
\begin{aligned}
\forall \ell: & \ M_{\ell} - \sum_{s_i\in C_{\ell}} z(i) + \sum_{b=1}^{m_{2,\ell}-1} 2^{b-1}y_{2,\ell,b} \\
&\quad + (|C_{\ell}| - M_{\ell} + 1 - 2^{m_{2,\ell}-1})y_{2,\ell,m_{2,\ell}} = 0. \label{eq:constraint5}
\end{aligned}
\end{equation}
Finally, for constraints (\ref{eq:constraint3}), we introduce slack binary variables $y_{3,j}$ to convert the inequality constraints to equality constraints:
\begin{equation}
\forall j: \sum_{\ell=1}^k x_{\ell}(j) + y_{3,j} -1 = 0. \label{eq:constraint6}
\end{equation}
We then can obtain the QUBO formulation by converting the equality constraints (\ref{eq:constraint4}) - (\ref{eq:constraint6}) to quadratic penalty terms and adding them to the objective function, the QUBO formulation is given as follows:

\begin{equation}
\begin{split}\label{eq:tmcdqubo}
\text{(QUBO)}\ \min &\ \sum_{\ell=1}^k\sum_{j\in T} x_{\ell}(j) + P \sum_{\ell=1}^k\sum_{v,w\in T}\frac{k_vk_w}{2|E|}x_{\ell}(v)x_{\ell}(w)\\
&+A\sum_{\ell=1}^k\sum_{s_i\in C_{\ell}}\bigg(z(i) - \sum_{j\in t_i} x_{\ell}(j) \\
&\quad + \sum_{b=1}^{m_{1,i}-1} 2^{b-1}y_{1,i,b}\\
&\quad + \bigg(|t_i|+1-2^{m_{1,i}-1})y_{1,i,m_{1,i}}\bigg)^2\\
&+B\sum_{\ell=1}^k\bigg(M_{\ell} - \sum_{s_i\in C_{\ell}} z(i) + \sum_{b=1}^{m_{2,\ell}-1} 2^{b-1}y_{2,\ell,b}\\
&\quad  + (M_{\ell} - |C_{\ell}| +1-2^{m_{2,\ell}-1})y_{2,\ell,m_{2,\ell}}\bigg)^2\\
&+C\sum_{j\in T}\bigg(\sum_{\ell=1}^k x_{\ell}(j) + y_{3,j} - 1\bigg)^2, \\
\text{s.t.} \ & \forall j, \forall \ell: x_{\ell}(j) \in \{0, 1\}, \ \forall i: z(i) \in \{0, 1\}, \\
&\forall \ell, \forall s_i\in C_{\ell}, \forall b: y_{1,i,b} \in \{0, 1\}, \\
&\forall \ell, \forall b: y_{2,\ell,b} \in \{0, 1\}, \forall j, y_{3,j} \in \{0, 1\},
\end{split}
\end{equation}
where $A, B, C > 0$ are parameters to penalize the violation of constraints.

\section{Empirical evaluation}
We present two test cases for the empirical evaluation. In Section \ref{sec:twitter}, the dataset (the election data) demonstrates results that are similar to the previous models tested on the same data. In addition, we further demonstrate the parametrization obstacle that is common to all models. In Section \ref{sec:pubmed}, the dataset (biomedical papers), we demonstrate the superiority of the proposed model on more challenging and less straightforward cluster data. Reproducibility materials: \href{https://anonymous.4open.science/r/explainability-cluster-descriptor-9307}{Link}

\subsection{USA Presidential Election 2016}\label{sec:twitter}
\paragraph{Dataset description} The experimental data set of Twitter was collected from 01/01/16 until 08/22/16 and covered the political primary season of the United States 2016 Presidential Election. The 880 most politically active Twitter users were chosen, and a graph $X$ was constructed based on the users' retweet behavior. That is, $X_{ij}$ (the weight of the directed edge $ij$) is the number of times user $i$ was retweeted by user $j$. We are given the result found by spectral clustering that divides $X$ into two clusters. The clustering algorithm found two natural communities, namely, pro-Democratic and pro-Republican. Also, the 136 most used political hashtags were collected to obtain dataset $Y$, which gives how many times each user tweeted with such hashtags.

We evaluate the QUBO formulation in Eq. (\ref{eq:tmcdqubo}) for TMCD and set different values for the parameters $M_{\ell}$ and $P$. The QUBO is solved using Fujitsu 3rd generation DA. Each instance solved on DA contained about 1000 binary variables and slack variables from the constraints. The time given to DA was limited to 30 seconds. The tags generated are presented in Tables \ref{tab:twitter1}-\ref{tab:twitter3}.

\begin{table*}
\centering
\caption{Tags generated with $M_1 = |C_1|=496, M_2 = |C_2| = 384$, that is, all users in the two clusters are covered. We experiment with the tag modularity parameter $P \in \{0, 1, 5\}$. When $P = 0$, the problem is equivalent to the QUBO of DTDM.}
\begin{tabularx}{0.815\textwidth}{*{8}{c}}
\toprule
\multicolumn{2}{c}{$P = 0$} & & \multicolumn{2}{c}{$P = 1$}  & & \multicolumn{2}{c}{$P = 5$}\\
Cluster 1 & Cluster 2 & & Cluster 1 & Cluster 2 & & Cluster 1 & Cluster 2 \\
\cline{1-2}
\cline{4-5}
\cline{7-8}
RNCinCLE	&	GOPdebate	& &	RNCinCLE	&	GOPdebate	& &	RNCinCLE	&	GOPdebate	\\
NeverTrump	&	Trump	& & 	NeverTrump	&	NYPrimary	& &	NeverTrump	&	Trump2016	\\
SuperTuesday	&	Trump2016	& &	SuperTuesday	&	NeverHillary	& &	SuperTuesday	&	CruzSexScandal	\\
DemDebate	&	NYPrimary	& &	IowaCaucus	&	CrookedHillary	& &	IowaCaucus	&	CrookedHillary	\\
IowaCaucus	&	IndianaPrimary	& &	DNCinPHL	&	ccot	& &	DNCinPHL	&	AmericaFirst	\\
DNCinPHL	&	CrookedHillary	& &	NHPrimary	&	AmericaFirst	& &	NHPrimary	&	NeverCruz	\\
Clinton	&	AmericaFirst	& &	TrumpTrain	&		& &	TrumpTrain	&		\\
trump	&		& &	Election2016	&		& &	Election2016	&		\\
TrumpTrain	&		& &	TrumpRally	&		& &	TrumpRally	&		\\
Election2016	&		& &	BernieSanders	&		& &	BernieSanders	&		\\
ImWithHer	&		& &	Hillary2016	&		& &	Hillary2016	&		\\
BernieSanders	&	&	&	USA	& &		&	USA	&		\\
TCOT	&	&	&	trump2016	&	&	&	trump2016	&		\\
USA	&		& &	UniteBlue	&		& &	UniteBlue	&		\\
trump2016	&	&	&	CCOT	&	&	&	CCOT	&		\\
	&		& &	BernieOrBust	&		& &	BernieOrBust	&		\\
\bottomrule
\end{tabularx}

\label{tab:twitter1}
\end{table*}

\begin{table*}
\centering
\caption{Tags generated with $M_1 = 471, M_2 = 365$, that is, we require about 95\% of the users to be covered. We experiment with the tag modularity parameter with $P \in \{0, 0.01, 0.1, 0.5, 1, 5\}$. When $P = 0$, the problem is equivalent to the QUBO of MinConCD.}
\begin{tabularx}{0.79\textwidth}{*{8}{c}}
\toprule
\multicolumn{2}{c}{$P = 0$} & & \multicolumn{2}{c}{$P = 0.01$} & & \multicolumn{2}{c}{$P = 0.1$}\\
Cluster 1 & Cluster 2 & & Cluster 1 & Cluster 2 & & Cluster 1 & Cluster 2 \\
\cline{1-2}
\cline{4-5}
\cline{7-8}
Trump	&	GOPdebate	& &	Trump	&	GOPdebate	& &	Trump	&	GOPdebate	\\
RNCinCLE	&	NeverTrump	& &	RNCinCLE	&	NeverTrump	& &	RNCinCLE	&	Trump2016	\\
SuperTuesday	&	Trump2016	& &	SuperTuesday	&	Trump2016	& &	NeverTrump	&	CruzSexScandal	\\
DemDebate	&	Clinton	& &	DemDebate	&	Clinton	& &	DemDebate	&	CrookedHillary	\\
IowaCaucus	&	&	&	IowaCaucus	&		&	& DNCleak	&	Republican	\\
\midrule
\multicolumn{2}{c}{$P = 0.5$} & & \multicolumn{2}{c}{$P = 1$} & & \multicolumn{2}{c}{$P = 5$}\\
Cluster 1 & Cluster 2 & & Cluster 1 & Cluster 2 & & Cluster 1 & Cluster 2 \\
\cline{1-2}
\cline{4-5}
\cline{7-8}
Trump	&	GOPdebate	& &	Trump	&	GOPdebate	& &	Trump	&	GOPdebate	\\
NeverTrump	&	Trump2016 &	&	NeverTrump	&	Trump2016	& &	NeverTrump	&	Trump2016	\\
SuperTuesday	&	BernieSanders	& &	SuperTuesday	&	BernieSanders	&	& SuperTuesday	&	BernieSanders	\\
DemDebate	&	NYC	& &	DemDebate	&	Republican	& &	DemDebate	&	Republican	\\

\bottomrule
\end{tabularx}

\label{tab:twitter2}
\end{table*}

\begin{table*}[htb]
\centering
\caption{Tags generated with $M_1 = 400, M_2 = 300$, that is, we require about 80\% of the users to be covered in both cluster. We experiment with the tag modularity parameter with $P \in \{0, 0.01, 0.1, 0.5, 1, 5\}$. When $P = 0$, the problem is equivalent to the QUBO of MinConCD.}
\begin{tabularx}{0.70\textwidth}{*{8}{c}}
\toprule
\multicolumn{2}{c}{$P = 0$} & & \multicolumn{2}{c}{$P = 0.01$} & & \multicolumn{2}{c}{$P = 0.1$}\\
Cluster 1 & Cluster 2 & & Cluster 1 & Cluster 2 & & Cluster 1 & Cluster 2 \\
\cline{1-2}
\cline{4-5}
\cline{7-8}
NeverTrump	&	GOPdebate	& &	RNCinCLE	&	GOPdebate	& &	Trump	&	GOPdebate	\\
SuperTuesday	&	Trump	& &	NeverTrump	&	Trump	& &	DemDebate	&	Trump2016	\\
DemDebate	&		& &	SuperTuesday	&		& &		&		\\
ImWithHer	&		& &	DemDebate	&		& & 		&		\\
\midrule
\multicolumn{2}{c}{$P = 0.5$} & & \multicolumn{2}{c}{$P = 1$} & & \multicolumn{2}{c}{$P = 5$}\\
Cluster 1 & Cluster 2 & & Cluster 1 & Cluster 2 & & Cluster 1 & Cluster 2 \\
\cline{1-2}
\cline{4-5}
\cline{7-8}
Trump	&	GOPdebate	& &	GOPdebate	&	Trump	& &	GOPdebate	&	Trump	\\
RNCinCLE	&	ImWithYou	& &		&		& &		&		\\
\bottomrule
\end{tabularx}

\label{tab:twitter3}
\end{table*}

\paragraph{Effect of Tag modularity} When the penalty parameter $P=0$, we obtain the special case of MinConCD. The tags generated from the algorithm provide us with some indications and features of the cluster they are associated with and can reveal some interesting information. As we increase the value of $P$, we add more weight on the tag modularity. Let $E_{t1} = \{(s_i, t)|\forall s_i\in C_1\}$ denote the set of edges that connect hashtag $t$ with the users in Cluster 1, and similarly, $E_{t2} = \{(s_i, t)|\forall s_i\in C_2\}$ is the set of edges that connect hashtag $t$ with the users in Cluster 2. To help us better understand the effect of tag modularity, we define \textbf{balance ratio (BR)} for each hashtag $t\in T$ as follows:
\begin{equation*}
\text{BR}_t = \frac{\min(|E_{t1}|, |E_{t2}|)}{\max(|E_{t1}|, |E_{t2}|)}.
\end{equation*}
If hashtag $t$ is connected with equal number of users from the two clusters then $\text{BR}_t = 1$ which means that $t$ is general to both clusters and not explaining cluster specific information. On the other hand, if hashtag $t$ is only connected with one cluster then, $\text{BR}_t = 0$. In this case, the hashtag should reveal more dedicated information about the cluster, which is more preferable regarding the explainability of that cluster. We compute the average of BR for the tags generated in Tables \ref{tab:twitter1}-\ref{tab:twitter2} for different $P$ values and present the results in Table \ref{tab:balanceratio}. As $P$ increases, the average BR decreases, which means that with more weight on the tag modularity, a preference to tags that contain cluster specific information over the general ones is given.

\begin{table*}
\centering
\caption{Average of BR of the tags generated in Tables \ref{tab:twitter1}-\ref{tab:twitter2}}
\begin{tabularx}{\textwidth}{cXXXXXX}
\toprule
&$P = 0$ & $P = 0.01$ & $P = 0.1$ & $P = 0.5$ & $P = 1$ & $P = 5$ \\
\midrule
Table 1: $M_1 = 496, M_2 = 384$ & 0.648 & 0.632 & 0.596 & 0.591 & 0.610 & 0.593 \\
\midrule
Table 2: $M_1 = 471, M_2 = 365$ & 0.921 & 0.921 & 0.694 & 0.840 & 0.840 & 0.840 \\
\bottomrule
\end{tabularx}

\label{tab:balanceratio}
\end{table*}

\paragraph{Sensitivity of $M_{\ell}$} Coverage of objects in explainable clustering is an important sensitive parameter that requires further investigation. When we enforce that all users have to be covered, that is, the special case of DTDM, shown in Table \ref{tab:twitter1}, we will need more tags to satisfy all constraints. However, with a relaxed coverage requirement, the number of tags we use is decreasing, as shown in Tables \ref{tab:twitter2}-\ref{tab:twitter3}. This makes the clustering more explainable. In Table \ref{tab:twitter3}, we present the results in which we relax the condition of covered explainable clustered objects and require to cover only about 80\% of them. While the behaviour of the model for $P=\{0,0.01,0.1 \}$ is still reliable, the increased penalty with $P=\{0.5,1,5\}$ does not find satisfactory explanation terms. In particular, it is important to understand the conditions under which \emph{no phase transition} occurs in the quality and number of explainability terms, i.e., there is no rapid change from the set of explaining to non-explaining terms in $T$.

\subsection{Biomedical Publications}\label{sec:pubmed}
To demonstrate the applicability of the TMCD model, we utilize it in  the natural language processing field, specifically, in the  literature-based discovery. To investigate the efficacy of our model, we compare it with the typical method used in the field, probabilistic topic modeling.  Literature-based discovery is one of the most rapidly growing fields in biomedical AI whose goal is to discover knowledge and generate scientific hypotheses using published scientific papers \cite{sybrandt2017moliere,sybrandt2020agatha}. Such types of analysis generate huge semantic ML models, and restricting the information domains with various types of topical modeling and clustering (such as the BioLDA \cite{wang2011finding}) is the most straightforward way to accelerate such systems. Given a collection of papers from PubMed and their categorical labels, we evaluate the model's ability to explain the topics behind the provided labels generated by the Latent Dirichlet Allocation (LDA). Overall, the LDA family of approaches is known to be extremely sensitive to the hyperparameters and convergence, so a proposed alternative approach for explainability is potentially of a broad impact.

\begin{table*}[htb]
\small
    \caption{Topics obtained with the LLDA and TMCD models, selected with the described hyperparamether tuning procedure, on PubMed dataset of infectious diseases. 
            Keywords produced by both models are shown in Overlap column. Scores are sorted based on their LLDA probabilities for a given topic where possible.
    }
    
    \begin{tabularx}{\textwidth}{lXXX}
        \toprule
        {} &                                                                                                                                                                                                             LLDA &                                                                                                                                                                                                                                        TMCD &                                                                                                                 Overlap \\
        \midrule
        HIV Infections &                          HIV-1, Anti-HIV Agents, Antiretro-viral Therapy, Highly Active, CD4 Lymphocyte Count, Viral Load, HIV, Sexual Behavior, Homosexuality, Male, CD4-Positive T-Lymphocytes, Sexual Partners &        HIV-1, Anti-HIV Agents, Antiretroviral Therapy, Highly Active, HIV, Sexual Behavior, Homosexuality, Male, Health Knowledge, Attitudes, Practice, Health Policy, Anti-Retroviral Agents, United States, Adult, Prevalence, Caregivers &                HIV-1, Anti-HIV Agents, Antiretroviral Therapy, Highly Active, HIV, Sexual Behavior, Homosexuality, Male \\
        \hline
        Measles        &                                                                               Measles Vaccine, Infant, Measles virus, Child, Preschool, Child, Disease Outbreaks, Vaccination, Antibodies, Viral, Rubella, Mumps &                                                                                                                    Measles Vaccine, Infant, Measles virus, Child, Disease Outbreaks, Antibodies, Viral, Mumps, Acute Disease, Poliomyelitis &                              Measles Vaccine, Infant, Measles virus, Child, Disease Outbreaks, Antibodies, Viral, Mumps \\
        \hline
        Hepatitis B    &  Hepatitis B virus, Hepatitis B Surface Antigens, Hepatitis B Antibodies, Hepatitis B Vaccines, DNA, Viral, Liver Neoplasms, Carcinoma, Hepatocellular, Hepatitis B Core Antigens, Hepatitis B e Antigens, Liver &                                                                                                                                                                                                                   Hepatitis B virus, Humans &                                                                                                       Hepatitis B virus \\
        \hline
        Rabies         &                                                                                   Animals, Rabies Vaccines, Rabies virus, Dogs, Mice, Antibodies, Viral, Dog Diseases, Vaccination, Bites and Stings, Chiroptera &                                                                                                                                                                                                      Animals, Rabies Vaccines, Rabies virus &                                                                                  Animals, Rabies Vaccines, Rabies virus \\
        \hline
        COVID-19       &                                                      SARS-CoV-2, Pandemics, Pneumonia, Viral, Coronavirus Infections, Betacoronavirus, COVID-19 Testing, Infection Control, China, Aged, 80 and over, Quarantine &                                                                                                                                                                       SARS-CoV-2, Pandemics, Cross-Sectional Studies, Retrospective Studies &                                                                                                   SARS-CoV-2, Pandemics \\
        \hline
        Hepatitis C    &                         Hepacivirus, Antiviral Agents, Hepatitis C Antibodies, RNA, Viral, Genotype, Interferon-alpha, Liver Cirrhosis, Recombinant Proteins, Substance Abuse, Intravenous, Hepatitis Antibodies &  Hepacivirus, Antiviral Agents, Hepatitis C Antibodies, Interferon-alpha, Liver Cirrhosis, Substance Abuse, Intravenous, Health Services Accessibility, Aged, Liver Neoplasms, Risk Factors, Hepatitis, Viral, Human, Liver Transplantation &  Hepacivirus, Antiviral Agents, Hepatitis C Antibodies, Interferon-alpha, Liver Cirrhosis, Substance Abuse, Intravenous \\
        \bottomrule
    \end{tabularx}

    \label{tab:llda_ising_overlap_kwds}
\end{table*}

\paragraph{Dataset description} Our source of information is PubMed, a large database of biomedical citations that include abstracts and other supporting information from various sources, containing more than 30 million entries. This data is publicly available via web API or FTP for large-scale processing. Each instance solved on DA contained about 17000 binary variables and slack variables from the constraints. The time given to DA was limited to 30 minutes.

Citations are supplied with additional metadata, such as manually curated Medical Subject Headings or MeSH Terms, which are used for PubMed indexing. These MeSH Terms can be used to represent a citation in a more compact and concise manner, significantly reducing the overall vocabulary size. We use MeSH for ground-truth clustering purposes, that is, if a citation contains a single specific keyword from a list of keywords defining clusters, this citation is considered belonging to this cluster. 

The list of keywords defining ground-truth clusters is based on widely known infectious diseases and includes the following 6 entries: \textit{COVID-19, HIV Infections, Hepatitis B, Hepatitis C,  Measles} and \textit{Rabies}. Based on the proposed strategy, we sample 500 PubMed citations for each cluster and use MeSH Terms to represent each document as a set of tags connected to it \emph{removing the MeSH Term used for clustering} (i.e., the ``trivial'' tag). We additionally filter out the tags that occurred less than 2 times within the entire dataset.

\paragraph{Topic modeling results} To demonstrate the TMCD model performance in topic modeling problem, we run the experiment with several modularity parameters $P$ (varying from 0 to 5), and set $M_{\ell} = 400, 1\leq \ell \leq 6$, that is, we require about 80\% of the citations to be covered. For comparison, we use a topic modeling approach called LLDA (Labeled Latent Dirichlet Allocation~\cite{ramage2009labeled_lda}), which is a more reliable and suitable for our purposes extension of the classic LDA model\cite{blei2003latent}. The main difference between LDA and LLDA models is that LLDA has an additional constraint on having 1-to-1 mapping between latent topics and provided document labels (that is the case with our sets $X$ and $Y$). 

The LLDA model is additionally optimized by hyperparameter tuning: $\alpha$ (represents the distribution of keywords over topics) and $\theta$ (represents the distribution of topics over documents) values form a 2d-grid and the optimization goal is to maximize the number of discriminative keywords in the provided topics, which do not occur in the common "background" topic representing all documents. All distributions of keywords over topics are limited to only top-$n$ keywords with the highest probabilities, where $n=10$.

For modularity parameters $P$ from the following list: [0.0, 0.0001, 0.0005, 0.001, 0.01, 0.1, 0.5, 1.0, 5.0], we determine the best modularity parameter $P$ for the TMCD model by performing a per-topic keyword overlap with the topics produced by LLDA. We maximize a ratio between the number of keywords shared with the LLDA model, and the overall number of tags produced by the TMCD model. We summarize the results obtained from TMCD with $P = 0.1$ and LLDA with $\alpha=10, \theta = 1$ in Table \ref{tab:llda_ising_overlap_kwds}.

In this experiment, we observe a significant difference between the LLDA and TMCD models. The QUBO optimization of TMCD prevented selecting more general and less explainable terms such as \emph{animals} and made it relevant only to the \emph{Rabies} cluster. Depending on the parametrization both approaches may exhibit different resolution scales of terms. However, it is easy to see that maximizing the modularity term in the TMCD model in combination with minimization of the total number of keywords  prevents choosing such terms as \emph{humans} in almost all cluster descriptors that do not contribute to explainability.

The main difference in the results comes from the distinct nature of the models: LLDA produces a probability distribution of the entire vocabulary of terms per topic (which we limit to top-$10$), whereas the TMCD model only outputs the terms themselves and does not allow cross-topical overlap. Clearly, both models' output contains relevant and noisy terms, but their overlap is (mostly) made of a small highly discriminative subset of the dataset vocabulary. \emph{Thus, the proposed scenario would be to use TMCD not instead, but in addition to traditional topic modeling to improve its performance and identify a handful of relevant terms. }

\section{Challenges and Deployment Obstacles}

\paragraph{Parameters Tuning} 
The QUBO formulation is unconstrained by construction, and the parameters on the penalty terms, i.e., $A, B, C >0$ in Eq. (\ref{eq:tmcdqubo}) will also affect the performance of the model. Typically, one would choose a large positive number to penalize the violation of constraints. The DA we used in this work supports automatically scaling of these parameters, however for most of the generic solvers, one might need to tune these parameters for optimal performance. Usually, such tuning is done by repetitive solving the model with different parameters in a more or less smart way, but it undoubtedly affects the complexity. Mitigating slow tuning requires further investigation.

\paragraph{Lack of reliable numeric explainability metric} Many recent works indicate that the limited number of ways to objectively compare explainability schemes represents an important problem in the field 
\cite{hoffman2018metrics,rosenfeld2021better}. However, the lack of quantitative methods to compare using secondary sets of tags (such as in our case) is even bolder. While human-based comparison as we can do with the demonstrated benchmarks is always illuminating, this approach is not scalable to larger datasets, numbers of tags and/or clusters. In particular, this is a critical problem when two explainability methods are different (e.g., topic modeling and our model). They will clearly not demonstrate comparable optimized objectives but this won't reveal much about the quality of explainability.

\paragraph{Noisy explainability terms}
Real-world data usually contains noise and the explainability problem is not an exception. In both applications we found terms making the task significantly more complicated. For example, in the Twitter dataset, there are hashtags frequently used by either party supporters (for example, “Trump” or “RNCinCLE”). They cover many users, but they do not describe the clusters distinctively. In the PubMed dataset we deal with the same problem, but, in addition, the number of MeSH terms used for citations annotation is quite large unlike the number of hashtags. However, the majority of these terms do not cover any sensible part of any cluster. While we filtered out terms that appeared only once and our approach takes care of high-frequency terms automatically, advanced strategies to filter out extremely low-frequency terms have to be developed to scale the proposed and other similar approaches.

\paragraph{Deployment challenges on acceleration hardware} 

The formulation of the tag descriptor problem as QUBO has many advantages due to the natural explainability of combinatorial objects. However, the inherent NP-hardness of the formulation creates computational challenges. In practice, real-word clustering problems with secondary tag sets can easily lead to very large-scale tag descriptor problems. In a post-Moore's law era, current quantum and quantum-inspired devices can only solve problems up to a limited number of variables. Whereas the third generation of the Digital Annealer is capable of solving problems up to 100,000 variables, which is still relatively small when working with real-world problems, devices such as the D-Wave quantum annealer scale to only about 200 variables\footnote{Note that the number of qubits is different from the number of variables that can effectively be solved.}, and other devices to at most a few thousand variables. \emph{Therefore, this is one of the biggest deployment challenges.} Another challenge is that such devices as the Digital Annealer are not exact solvers, i.e., the optimal solution is not guaranteed. This leads to the difficulties associated with parameter tuning. Lastly, there is a challenge associated with the precision of encoding the problems onto the hardware. Whereas the Digital Annealer utilizes digital circuits to encode a problem and can easily scale to its current 64-bit precision, many of the other quantum and quantum-inspired devices are analog devices that only encode problems with a very limited number of bits, usually less than 5 bits. This greatly affects the type of problems that fully utilize the power of these devices. That said, generalized global solvers perform much worse than specialized hardware even on many small instances.

\section{Conclusions}
In this work, we propose an explainability model, the tag modularity cluster descriptor, that makes the previous cluster descriptor models more practical on real-life datasets with nonuniform cluster tag distributions. We introduce its QUBO formulation and explore the possibility of utilizing specialized hardware to solve such problems. We demonstrate experiments by solving the problems using Fujitsu Digital Annealer and use the Twitter and PubMed datasets to validate the approach. The QUBO formulation of the model makes it naturally explainable and tunable in contrast to such explainability approaches as topical modeling which we compare with. Fully scalable deployment of the proposed approach requires further investigation that combines optimization model with hardware acceleration in such directions as parameter tuning, introducing advanced objective quality metric for explainability, and carefully addressing computational precision requirements.

\bibliographystyle{IEEEtran}
\bibliography{bib}
\end{document}